\documentclass[10pt,twocolumn,letterpaper]{article}

\usepackage{cvpr}              

\usepackage{cuted}
\usepackage{capt-of}
\usepackage{booktabs}
\usepackage{array}
\usepackage{tabularx}

\newcolumntype{L}[1]{>{\raggedright\arraybackslash}p{#1}}

\usepackage{multirow}
\usepackage{booktabs}
\usepackage{booktabs}
\usepackage{tabularx}
\usepackage{multirow}
\usepackage{placeins} 

\usepackage{tcolorbox}
\tcbuselibrary{breakable}

\newtcolorbox{promptbox}[1][]{
  breakable,
  colback=gray!8,
  colframe=gray!45,
  fonttitle=\bfseries\small\sffamily,
  title=#1,
  left=4pt, right=4pt, top=3pt, bottom=3pt,
  fontupper=\small\ttfamily
}

\definecolor{cvprblue}{rgb}{0.21,0.49,0.74}
\usepackage[breaklinks,colorlinks,allcolors=cvprblue]{hyperref}

\def\paperID{*****}
\def\confName{CVPR}
\def\confYear{2026}

\newcommand{\papertitle}{Privacy-Aware Synthetic Video Benchmarking and Relational Evaluation for Worker-Under-Suspended-Load Detection}

\title{\papertitle}

\author{
Anshu Singh\thanks{Both authors contributed equally to this work.}
\quad
Alejandro Seif\footnotemark[1]\\
Government Technology Agency of Singapore\\
{\tt\small \{singh\_anshu, alejandro\_seif\}@tech.gov.sg}
}

\begin{document}

\maketitle

\begingroup
\renewcommand{\thefootnote}{}
\footnotetext{Accepted to the 3rd Workshop on Synthetic Data for Computer Vision (SynData4CV), CVPR 2026.}
\endgroup


    
    

\begin{abstract}
Publicly shareable construction-video benchmarks remain scarce, especially for safety-critical hazards that are rare, dangerous to stage, and difficult to release. We study \emph{worker under suspended load}, a relational hazard that depends on worker--load geometry and temporal persistence rather than object detection alone. We introduce \textit{SynthSite}, a focused synthetic video benchmark of 55 clips spanning varied load configurations, viewpoints, clutter, occlusions, and surveillance conditions, together with a privacy-aware hybrid generation workflow that supports both publicly shareable benchmark creation and privacy-constrained synthetic video generation.

We then ask whether worker appearance can be suppressed without undermining downstream hazard recognition. Under five whole-body privacy conditions, we evaluate worker and load retention, localization stability, and clip-level hazard recognition. We find that structure-preserving obfuscations retain substantially more downstream utility than appearance-smoothing baselines, and that preserving a raw visual reference alone does not guarantee the strongest agreement with human hazard labels. These findings suggest that privacy evaluation for construction safety analytics should assess not only appearance suppression, but also preservation of the geometric cues required for hazard reasoning. Our dataset and code are available at \text{https://huggingface.co/datasets/govtech/SynthSite}
\end{abstract}
\begin{figure*}[t]
  \centering
  \includegraphics[width=\linewidth]{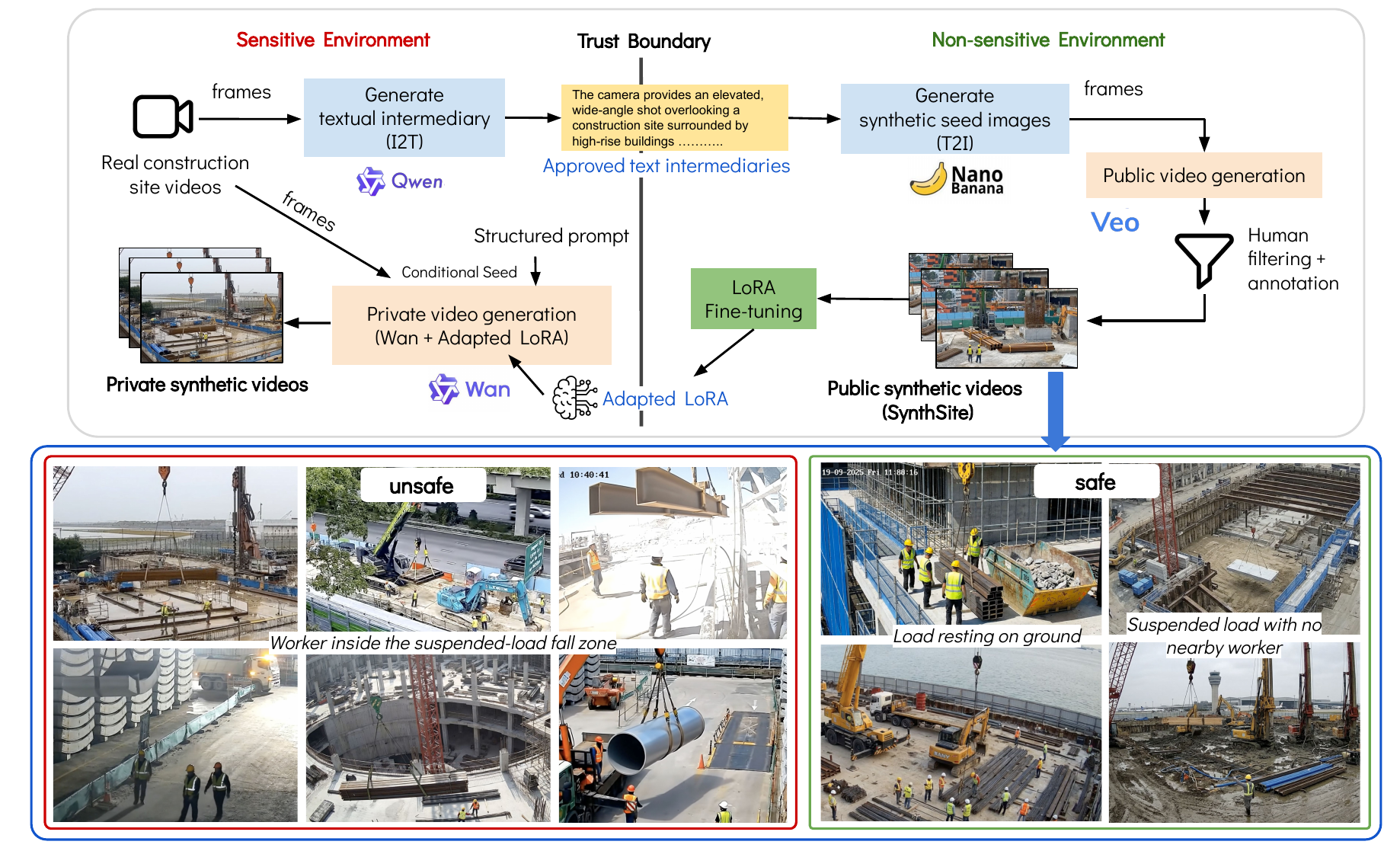}
\caption{\textbf{Privacy-aware hybrid workflow and representative \textsf{SynthSite} clips.} Raw construction footage remains inside the sensitive environment, where a local vision--language model (Qwen2.5-VL) generates approved textual intermediaries. These intermediaries cross into the non-sensitive environment, where Nano Banana 2 produces synthetic seed images and Veo 3.1 Fast generates candidate publicly shareable synthetic videos. Curated and annotated public videos form \textsf{SynthSite}, while a subset is used to train a LoRA that is imported back into the sensitive environment and combined with Wan2.2-I2V-A14B, structured prompts, trigger words, and optional local reference frames to generate privacy-constrained internal synthetic videos. The example clips illustrate both unsafe cases, where workers occupy the suspended-load fall zone, and safe cases, where loads are grounded or workers remain outside the fall zone, across diverse machinery, viewpoints, clutter, occlusion, and surveillance conditions.}
  \label{fig:synthetic_mosaic}
\end{figure*}

\section{Introduction}
\label{sec:intro}

Construction sites remain among the most hazardous work environments, with safety-critical incidents arising from heavy machinery, dynamic layouts, and changing worker--equipment interactions. Safety regulations therefore enforce exclusion zones around dangerous operations, including suspended loads and their fall zones~\cite{osha1926loadclear}. Recent reports continue to document fatal injuries involving workers struck by falling or moving objects in construction and related sectors~\cite{mom2024wshreport,wshc2024struckby}. These risks make automated safety monitoring an important application area for computer vision.

Many practically important construction hazards, however, are not object categories but \emph{relations} that unfold over time. A worker may enter a machine danger zone, equipment may transition into an unsafe state, or a suspended load may become hazardous only relative to nearby workers. We study one such hazard: \emph{worker under suspended load}, where risk arises when a worker is located within the fall zone beneath a suspended load.

This setting is challenging for two reasons. First, it is inherently \emph{relational} and \emph{spatiotemporal}: a system must localize workers and loads, infer whether a load is suspended, estimate an approximate fall zone, and aggregate evidence over time under clutter, occlusion, and unstable surveillance imagery. Prior work has shown that crane load fall-zone monitoring is feasible~\cite{chian2022dynamic,pazari2023enhancing}, while also highlighting the difficulty of robust suspended-load detection under wide appearance variation~\cite{chian2022dynamic}. Second, publicly shareable \emph{video} benchmarks for this hazard are scarce. In our setting, available real construction CCTV did not provide enough positive suspended-load events or controlled diversity for benchmark construction, and operational footage is often difficult to release because it may expose workers, site layouts, equipment configurations, and other sensitive details.

In this setting, synthetic video is not just a data augmentation tool; it is a practical mechanism for constructing shareable benchmarks for rare and safety-critical events. Recent work on sensitive-data synthesis with foundation models further suggests that privacy-aware generation is shaped not only by model capability, but also by deployment, infrastructure, and model-access constraints~\cite{wang2025spti,brito2025synthguard,reynaud2024echonet}.

In this work, we introduce \textsf{SynthSite}, a focused synthetic video benchmark for \emph{worker under suspended load}. \textsf{SynthSite} contains 55 clips spanning varied suspended materials, viewpoints, clutter, occlusions, and surveillance conditions relevant to construction monitoring. To support benchmark creation, we develop a \textbf{privacy-aware hybrid generation workflow} that keeps raw construction footage inside a trusted environment while allowing only approved textual intermediaries to cross the trust boundary for downstream synthesis. This workflow supports both publicly shareable benchmark videos and privacy-constrained internal synthetic video generation.

Beyond benchmark construction, we study a practical question for privacy-conscious safety analytics: \emph{how much worker appearance can be suppressed before relational hazard recognition degrades?} Semantically, worker identity is not required for this task, since hazard reasoning depends on worker--load geometry rather than personal identity. In practice, however, visual obfuscation can still degrade worker retention and geometric stability. We therefore define a \textbf{relational privacy evaluation} under whole-body worker obfuscation, measuring clip-level hazard recognition together with worker retention, load retention, and localization stability.

Our contributions are threefold:
\begin{itemize}
\item We introduce \textsf{SynthSite}, a focused synthetic construction video benchmark for \emph{worker under suspended load}, a safety-critical relational hazard that is difficult to capture, rare in available real CCTV, and underrepresented in public construction-vision benchmarks.

\item We propose a \textbf{privacy-aware hybrid generation workflow} that separates sensitive and non-sensitive synthesis stages using approved textual intermediaries, enabling publicly shareable benchmark creation and privacy-constrained internal synthetic video generation under realistic deployment constraints.

\item We present a \textbf{relational privacy evaluation} for construction safety analytics, showing how whole-body worker obfuscation affects downstream hazard reasoning, and showing that agreement with the raw pipeline and agreement with human hazard labels are not identical objectives.
\end{itemize}

\section{Related Work}
\label{sec:related_work}

\paragraph{Construction safety benchmarks and suspended-load monitoring.}
Public construction-safety benchmarks remain dominated by object-centric image datasets such as MOCS~\cite{xue2021mocs} and SODA~\cite{duan2022soda}. More recent video and simulation efforts move toward dynamic monitoring, but primarily target PPE compliance, vehicle motion, or general proximity events rather than suspended-load hazards~\cite{shrigandhi2025csod24,wu2024consynth}. Prior work on crane load fall-zone monitoring shows that suspended-load reasoning is feasible but challenging under clutter, occlusion, and wide variation in load appearance~\cite{chian2022dynamic,pazari2023enhancing}. Relative to these lines of work, \textsf{SynthSite} focuses specifically on \emph{worker under suspended load}, a relational video setting that depends on suspension-state inference, worker--load geometry, and temporal reasoning.

\paragraph{Privacy-aware synthesis workflows for sensitive video.}
Recent work has framed sensitive-data synthesis as a workflow problem shaped not only by model capability, but also by infrastructure, policy, and model-access constraints~\cite{brito2025synthguard,wang2025spti}. Release-oriented medical-video synthesis further illustrates the need to control what leaves the sensitive environment~\cite{reynaud2024echonet}. Our work follows this line by keeping raw construction footage inside a trusted environment and exporting only approved textual intermediaries for downstream generation, enabling benchmark creation without directly releasing operational sensitive video.

\paragraph{Privacy-preserving video analytics.}
Visual anonymization is a privacy--utility trade-off: suppressing appearance can also distort the cues required for downstream understanding~\cite{deconinck2024visual,hukkelas2023deepprivacy2}. Recent work in privacy-preserving video understanding likewise shows that anonymization can degrade spatiotemporal reasoning~\cite{fioresi2025splavu,chen2025stegavar}. We study this issue in a largely unexplored construction setting, asking whether whole-body worker obfuscation can hide identity while preserving the worker--load geometry needed for relational hazard detection.

\begin{table}[t]
\centering
\caption{Scene-level diversity of the 55 clips in \textsf{SynthSite}, estimated automatically from the first frame of each video using Gemini~2.5 Flash.}
\label{tab:clip-diversity}
\small
\setlength{\tabcolsep}{5pt}
\begin{tabular}{llcc}
\toprule
\textbf{Attribute} & \textbf{Category} & \textbf{Count} & \textbf{Percent (\%)} \\
\midrule
Proximity & Far         &  7 & 13 \\
          & Mid         & 40 & 73 \\
          & Close       &  8 & 14 \\
\midrule
Workers   & 1           &  8 & 15 \\
          & 2--5        & 27 & 49 \\
          & 6+          & 20 & 36 \\
\midrule
Machinery & Earthmoving &  1 &  2 \\
          & Lifting     & 32 & 58 \\
          & Mixed       & 22 & 40 \\
\midrule
Lighting  & Bright      & 28 & 51 \\
          & Overcast    & 26 & 47 \\
          & Dim/Night   &  1 &  2 \\
\midrule
Scene     & Open site   & 45 & 82 \\
          & Scaffold    &  9 & 16 \\
          & Roadwork    &  1 &  2 \\
\bottomrule
\end{tabular}
\end{table}

\section{\textsf{SynthSite}: Synthetic Worker-Under-Suspended-Load Video Dataset}

\subsection{Dataset Overview}

We introduce \textsf{SynthSite}, a focused synthetic video dataset for the safety-critical construction hazard of \emph{worker under suspended load}. The dataset contains 55 clips, with 28 \emph{safe} and 27 \emph{unsafe} videos. Each clip is 5--10 seconds long, generated at 15 FPS with a native resolution of $1280\times720$.

\textsf{SynthSite} is designed to capture variation relevant to construction surveillance and relational hazard reasoning. It includes diverse heavy machinery and suspended materials, such as steel I-beams, precast tunnel segments, and rebar cages, together with practical CCTV-like challenges including illumination changes, wide viewpoints, clutter, scale variation, and partial occlusion (Figure~\ref{fig:synthetic_mosaic}). As summarized in Table~\ref{tab:clip-diversity}, the clips vary in worker--load proximity, worker density, machinery context, lighting, and scene type. We position \textsf{SynthSite} as a focused benchmark for a narrowly defined relational hazard rather than as a large-scale surveillance corpus.

Because \emph{worker under suspended load} is a relational safety label rather than a simple object category, benchmark construction required manual quality control and rubric-guided annotation (Sections~\ref{sec:syn_quality} and~\ref{sec:syn_annotations}). We therefore prioritize controlled curation over scale.


\subsection{Privacy-Aware Hybrid Video Generation Workflow}
\label{sec:private_generation}

Real construction surveillance footage is often operationally sensitive, while the highest-fidelity video generators are typically closed-source and externally hosted. This creates a practical tension: strong generation quality is desirable, but raw footage cannot be moved freely across trust boundaries. We therefore design a \emph{privacy-aware hybrid video generation workflow} that supports both publicly shareable benchmark creation and internally controlled synthetic video generation (Figure~\ref{fig:synthetic_mosaic}).

We use the term \emph{privacy-aware} in an operational rather than formal sense. Raw footage remains inside a trusted environment, and only approved textual intermediaries are allowed to cross the boundary. In our instantiation, these intermediaries are structured scene descriptions or reverse prompts that summarize hazard-relevant scene composition, worker--equipment relations, and environmental context (Appendix~\ref{app:reverse_prompt_examples}). We do not claim a formal privacy guarantee; rather, the workflow is designed to reduce direct exposure of sensitive visual data while preserving sufficient structure for downstream synthesis~\cite{brito2025synthguard,reynaud2024echonet,wang2025spti}.

In the configuration used in this paper, raw frames remain in the \textbf{sensitive environment}, where a locally deployed vision--language model converts them into approved textual intermediaries. These are transferred to the \textbf{non-sensitive environment}, where external image and video generators produce candidate publicly shareable synthetic videos. After human filtering and annotation, a curated subset of these videos forms \textsf{SynthSite}. The same shareable synthetic videos are also used to adapt a locally deployable open video model via LoRA~\cite{hu2021lora}, after which the adapted model is imported back into the sensitive environment for internally controlled generation.

For the benchmark-generation branch, each candidate clip is guided by two complementary conditioning signals: a synthetic seed image, which anchors scene layout and coarse geometry, and a structured hazard prompt, which specifies suspended objects, worker placement, viewpoint, and nuisance factors such as clutter, occlusion, and illumination. For the internal-generation branch, the adapted local model uses structured prompts together with optional local reference frames that remain inside the sensitive environment. Because \emph{worker under suspended load} is a relational and spatiotemporal hazard, generation is treated as a guided proposal process followed by human filtering (Section~\ref{sec:syn_quality}) and annotation, rather than as a one-shot reliable synthesis mechanism.

Detailed implementation choices, including model allocation across environments, textual intermediary design, compute setup, and LoRA adaptation settings, are provided in Appendix~\ref{app:generation_details}.


\subsection{Quality Control}
\label{sec:syn_quality}

Before hazard annotation, all generated clips are manually screened for generation failures that would invalidate downstream labeling or relational hazard reasoning. Roughly half of the generated candidate clips were discarded during this stage. We discard clips containing severe synthesis failures, including hallucinated artifacts, duplication or fusion of key entities, abrupt appearance or disappearance of workers, loads, or machinery, temporally inconsistent depiction of workers or loads, physically implausible motion, or severe temporal corruption such as abrupt scene transitions.

We retain minor imperfections that do not prevent reliable hazard interpretation, including mild geometric distortion, low resolution, corrupted text, limited worker motion, and occasional dropped or inconsistent frames. This is intentional: the goal is not photorealistic perfection, but surveillance-style clips in which the worker--load relationship remains visually interpretable. Quality control is therefore applied as a task-oriented filtering stage, excluding clips that break spatial or temporal plausibility while retaining artifacts that do not invalidate the safe/unsafe label.

\subsection{Annotations}
\label{sec:syn_annotations}
Each clip is assigned a binary safety label (\emph{safe}/\emph{unsafe}) using a rubric-based event-occurrence rule and custom annotation platform, both described in Appendix~\ref{app:rubric}. A clip is labeled \emph{unsafe} when a visibly suspended load is present and a worker remains within the fall zone beneath the load for at least 1 second (15 consecutive frames); otherwise, it is labeled \emph{safe}. The rubric also defines common hard negatives, including clips where (i) no worker is present within the fall zone, (ii) the load is not physically suspended, or (iii) the scene creates a misleading impression of suspension due to perspective or occlusion. Clips with severe perspective ambiguity, where the worker--load relationship cannot be determined reliably, are excluded.

Annotations were collected from a pool of eight annotators briefed using a construction-safety-informed rubric. Each clip was independently reviewed by two annotators using the custom annotation platform, and disagreements were resolved by a third annotator. Cohen's $\kappa$ was 0.308. Because the task requires fine-grained spatial reasoning under clutter, occlusion, and perspective distortion, we rely on adjudication and exclusion of unresolved ambiguous clips to improve label reliability.
\begin{figure*}[t]
  \centering
  \includegraphics[width=\linewidth]{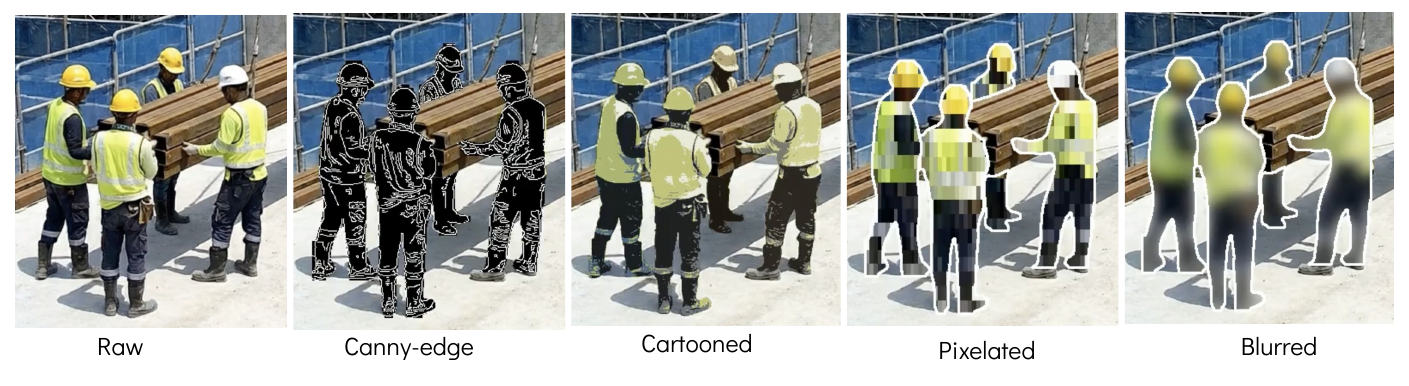}
\caption{\textbf{Relational privacy evaluation under worker obfuscation.} The same worker-under-suspended-load scene is shown under five evaluation conditions: a raw-video reference and four whole-body obfuscation settings: Canny-edge, cartooned, pixelated, and blurred worker representations.}
  \label{fig:privacy_relational_overview}
\end{figure*}

\section{Relational Privacy Evaluation}
\label{sec:method_relational_privacy}

We study whether \emph{worker under suspended load} can be recognized while suppressing worker-identifying appearance cues. Under our annotation rubric (Appendix~\ref{app:rubric}), a clip is labeled \emph{unsafe} when a visibly suspended load is present and a worker remains within the fall zone beneath the load for at least 1 second (15 consecutive frames); otherwise, it is labeled \emph{safe}. The task is therefore inherently \emph{relational} and \emph{spatiotemporal}: prediction depends on the worker--load relationship over time rather than on worker identity.

\subsection{Whole-Body Obfuscation Conditions}
\label{sec:whole_body_obfuscation}

We evaluate five conditions: a \textbf{raw-video reference} and four whole-body obfuscation settings: \textbf{Canny-edge body}, \textbf{cartooned body}, \textbf{pixelated body}, and \textbf{blurred body} (Figure~\ref{fig:privacy_relational_overview}). We use whole-body obfuscation rather than face-only redaction because worker faces are typically small, oblique, or occluded in construction surveillance. In such settings, identity-related cues may still be conveyed through soft biometrics such as clothing, build, and gait~\cite{zheng2016personreid,hukkelas2023deepprivacy2}.

These transformations span practically relevant anonymization families with different privacy--utility trade-offs~\cite{deconinck2024visual}. Canny-edge and cartooning apply stronger appearance abstraction while preserving coarse body structure, whereas pixelation and blur serve as standard visual anonymization baselines~\cite{wu2018privacyaction,erdelyi2017privacy}. All four are designed to preserve worker extent and position sufficiently for instance-level localization, which is essential because hazard recognition depends on worker--load geometry rather than photometric identity. To ensure a fair comparison, all obfuscations are applied only to worker regions obtained from a shared person-segmentation stage, with identical masks reused across methods; implementation details are provided in Appendix~\ref{app:obfuscation_details}.

Although the benchmark is synthetic, worker obfuscation remains relevant as an additional privacy layer. Synthetic visual data are not automatically privacy-safe: prior work has highlighted disclosure risks in synthetic-data release and memorization risks in high-capacity generative models~\cite{stadler2022synthetic,carlini2023extracting}. We therefore study whether appearance can be suppressed further while preserving the relational cues needed for hazard recognition.


\subsection{Relational Hazard Detection Pipeline}
\label{sec:relational_hazard_pipeline}

We formulate \emph{worker under suspended load} as a geometry-driven relational detection problem. Rather than attempting full 3D physical reconstruction from monocular surveillance video, we use a practical 2D proxy based on detections, motion cues, and temporal persistence. Given a video clip $\mathcal{V}=\{I_t\}_{t=1}^T$, the pipeline has three stages: (i) detect and track workers and hoisted loads, (ii) infer frame-level worker--load risk relations, and (iii) aggregate these into a clip-level safe/unsafe decision. Full implementation details are provided in Appendix~\ref{app:relational_hazard_pipeline_details}.

\paragraph{Detection and suspension inference.}
For each frame $I_t$, we detect workers and hoisted loads using YOLO-World~\cite{cheng2024yoloworld} and associate detections over time with ByteTrack~\cite{zhang2022bytetrack}. We use an open-vocabulary detector because task-specific lifted-load annotations are scarce, making supervised detector training outside the scope of this study. In a small pilot, YOLO-World was more suitable than Grounding DINO for this recall-sensitive surveillance setting in both usable lifted-load detections and inference latency; Appendix~\ref{app:detector_selection} provides prompt-design and detector-selection details.

For each worker, we define the worker footpoint as the bottom-center of the bounding box, denoted by $(x^{\mathrm{foot}}_{\mathrm{worker}}, y^{\mathrm{foot}}_{\mathrm{worker}})$. For each load, we use $x_{\mathrm{load}}$ for the horizontal center of the bounding box and $y^{\mathrm{bottom}}_{\mathrm{load}}$ for its bottom edge. Image coordinates increase downward in the vertical direction. To distinguish a suspended load from one resting on the ground, we use nearby workers as local ground proxies. For a load with width $w_{\mathrm{load}}$ and fixed horizontal expansion factor $\alpha$, only workers whose footpoints satisfy
\begin{equation}
|x^{\mathrm{foot}}_{\mathrm{worker}} - x_{\mathrm{load}}| < \alpha \, w_{\mathrm{load}}
\label{eq:nearby_worker}
\end{equation}
are considered. For each such worker--load pair, we compute a normalized clearance ratio
\begin{equation}
r^{\mathrm{clr}} =
\frac{y^{\mathrm{foot}}_{\mathrm{worker}} - y^{\mathrm{bottom}}_{\mathrm{load}}}{h_{\mathrm{worker}}},
\label{eq:clearance_ratio}
\end{equation}
where $h_{\mathrm{worker}}$ is the detected worker height in pixels. A load is treated as suspended when the maximum nearby-worker clearance exceeds a threshold. Normalizing by worker height improves robustness to viewpoint and scale variation. This suspension proxy is intentionally local and conservative, and is less reliable when no suitable nearby worker is available.

\paragraph{Fall-zone proxy and temporal decision.}
Once a load is inferred to be suspended, we estimate a trapezoidal fall-zone proxy beneath it. The region widens from the load toward the ground to reflect increasing lateral displacement during a possible fall. Its extent is computed from an estimate of load height above ground, recent horizontal motion, and a minimum drift floor that accounts for residual crane swing. A worker is considered exposed when the worker footpoint lies inside this trapezoidal region, subject to a vertical compatibility gate.

For each frame, we define a binary hazard indicator
\begin{equation}
h_t =
\mathbb{I}\!\left(
\begin{aligned}
\exists\, i,j \;\text{s.t. } &\text{load } j \text{ is suspended} \\
&\text{and worker } i \text{ lies in its fall-zone proxy}
\end{aligned}
\right)
\label{eq:frame_hazard}
\end{equation}
where $\mathbb{I}(\cdot)$ is the indicator function. We then aggregate $h_t$ over a 1-second sliding window and label a clip as \emph{unsafe} when sufficiently persistent hazardous evidence is observed within the window. To remain robust to transient detector dropouts, this temporal decision is a relaxed proxy rather than a strict enforcement of 15 consecutive positive frames; small negative gaps inside otherwise positive runs may be filled before counting. This proxy provides a lightweight basis for evaluating whether worker obfuscation preserves the relational cues needed for downstream hazard recognition.


\begin{table*}[t]
\centering
\caption{Effect of worker obfuscation on relational hazard detection in \textsf{SynthSite}. Columns under \emph{Raw-reference stability} measure consistency relative to the raw-video reference pipeline; columns under \emph{Agreement with human annotations} measure clip-level hazard recognition against human safe/unsafe labels. Higher is better for $F_2$, worker retention, and load retention; lower is better for worker and load localization jitter.}
\label{tab:privacy_results_combined}
\small
\setlength{\tabcolsep}{4pt}
\begin{tabular}{lccccc|c}
\toprule
& \multicolumn{5}{c|}{\textbf{Raw-reference stability}} & \multicolumn{1}{c}{\textbf{Agreement with human annotations}} \\
\cmidrule(r){2-6} \cmidrule(l){7-7}
\textbf{Method} & \textbf{$F_2 \uparrow$} & \textbf{Worker Ret. $\uparrow$} & \textbf{Load Ret. $\uparrow$} & \textbf{Worker Jitter $\downarrow$} & \textbf{Load Jitter $\downarrow$} & \textbf{$F_2 \uparrow$} \\
\midrule
RAW         & Ref.           & Ref.            & Ref.            & Ref.            & Ref.            & 0.757 \\
Canny-edge  & \textbf{0.964} & 80.2\%          & 85.2\%          & 2.17\%          & 1.48\%          & 0.742 \\
Cartooning  & 0.963          & \textbf{88.9\%} & 86.1\%          & \textbf{1.35\%} & \textbf{1.03\%} & \textbf{0.767} \\
Pixelation  & 0.913          & 63.1\%          & \textbf{86.8\%} & 1.82\%          & 1.45\%          & 0.762 \\
Blur        & 0.876          & 41.4\%          & 77.2\%          & 1.44\%          & 2.31\%          & 0.671 \\
\bottomrule
\end{tabular}
\end{table*}

\begin{figure*}[t]
  \centering
  \includegraphics[width=0.90\linewidth]{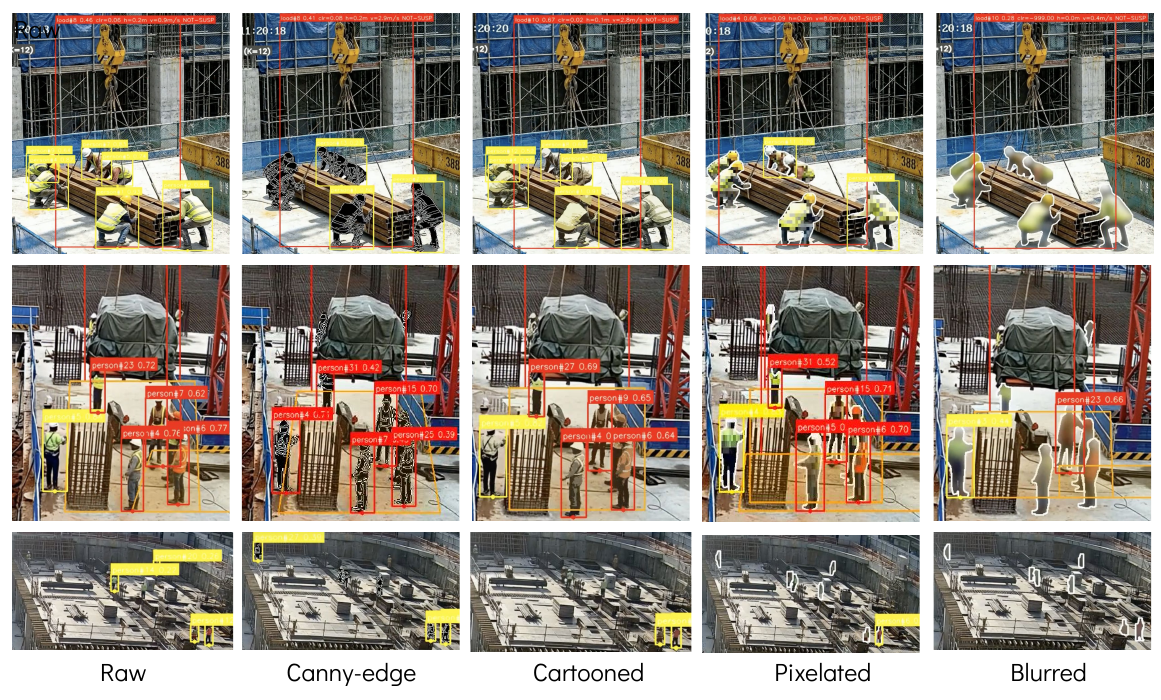}

\caption{\textbf{Qualitative examples from relational privacy evaluation under whole-body worker obfuscation.} Each row shows the same scene under five conditions: raw, Canny-edge, cartooned, pixelated, and blurred worker representations. Orange overlays denote the estimated fall-zone proxy; red boxes indicate unsafe worker--load detections, and yellow boxes indicate safe workers. Consistent with the quantitative results, structure-preserving obfuscations, especially cartooning and Canny-edge, better preserve worker localization and downstream relational hazard reasoning than pixelation and blur. The middle rows illustrate cross-entity effects, where worker obfuscation perturbs load localization and fall-zone geometry through the shared detection pipeline. The bottom row shows a distant-view failure case, where small worker scale degrades relational hazard inference across conditions.}
  
  \label{fig:obfuscation_results_qualitative}
\end{figure*}

\subsection{Evaluation Setup and Metrics}
\label{sec:relational_eval_protocol}

We evaluate the 55 clips in \textsf{SynthSite} from two complementary perspectives: (\textit{i}) \emph{raw-reference stability}, i.e., whether worker obfuscation preserves the behavior of the relational hazard pipeline relative to the raw-video reference, and (\textit{ii}) \emph{agreement with human annotations}, i.e., whether clip-level hazard predictions remain aligned with the human safe/unsafe labels. The same relational hazard pipeline and decision thresholds are used for all privacy conditions; only the video input is changed.

\paragraph{Raw-reference stability.}
We measure \textbf{worker retention} and \textbf{load retention} by matching detections in each obfuscated video to detections in the corresponding raw video using an IoU-based criterion. These metrics quantify how well the two detection primitives required for downstream hazard reasoning are preserved under obfuscation. We also measure \textbf{worker localization jitter} and \textbf{load localization jitter} to quantify the geometric stability of matched detections:
\begin{equation}
\mathrm{Jitter} =
\frac{\lVert c_{\mathrm{raw}} - c_{\mathrm{priv}} \rVert_2}{d_{\mathrm{raw}}},
\label{eq:jitter}
\end{equation}
where $c_{\mathrm{raw}}$ and $c_{\mathrm{priv}}$ are the center coordinates of matched bounding boxes in the raw and obfuscated settings, respectively, and $d_{\mathrm{raw}}$ is the diagonal length of the raw bounding box. Finally, we report clip-level \textbf{$F_2$} against the raw-video predictions, which measures hazard-decision consistency under privacy transformations. Raw-reference stability therefore reflects consistency with the raw proxy, not correctness with respect to the human annotations.

\paragraph{Agreement with human annotations.}
We separately evaluate clip-level predictions against the human safe/unsafe annotations, treating \emph{unsafe} as the positive class. Because this is a recall-sensitive safety setting, we emphasize clip-level \textbf{$F_2$} as the primary metric.



\subsection{Results and Discussion}
\label{sec:results_discussion}

Table~\ref{tab:privacy_results_combined} summarizes raw-reference stability and agreement with human annotations, while Figure~\ref{fig:obfuscation_results_qualitative} shows representative examples. Overall, cartooning yields the strongest agreement with human annotations ($F_2=0.767$), whereas Canny-edge gives the highest raw-reference stability ($F_2=0.964$). This distinction is important: preserving the behavior of the raw pipeline is not identical to preserving the strongest human-aligned hazard decisions.

\paragraph{The primary degradation is reduced worker retention, not geometric drift.}
Once detections are retained, their geometry remains relatively stable across privacy settings. Worker localization jitter stays within 1.35\%--2.17\% of the raw bounding-box diagonal, and load localization jitter within 1.03\%--2.31\%. By contrast, worker retention varies much more strongly, from 88.9\% for cartooning to 41.4\% for blur. This indicates that the main effect of obfuscation is not large localization error, but reduced \emph{worker detectability}. In other words, the relational hazard logic remains largely intact once a worker is still detected; the dominant failure mode is that some privacy transformations make the worker harder to retain as a stable instance.

\paragraph{Structure-preserving obfuscations best preserve relational hazard cues.}
Blur likely performs worst because it suppresses local contrast, articulation, and silhouette detail, reducing instance separability and stable worker retention. Cartooning likely performs best because it suppresses appearance texture while preserving coherent human structure, yielding the highest worker retention and the strongest agreement with human annotations. Pixelation is also informative: although its worker retention is substantially lower than cartooning, its human-annotation $F_2$ remains close. This suggests that clip-level unsafe classification depends on preserving sufficient relational evidence over time rather than perfectly retaining every worker instance. Overall, the results suggest that, for \emph{worker under suspended load}, preserving worker geometry matters more than preserving photorealistic appearance.

\paragraph{Worker obfuscation perturbs the load channel, but less than the worker channel.}
Although the load itself is never obfuscated, load detection still varies across privacy settings: load retention ranges from 77.2\% to 86.8\%, and load localization jitter from 1.03\% to 2.31\%. One possible explanation is that the open-vocabulary detector is sensitive not only to the load region itself but also to surrounding context such as nearby workers, scale, and scene layout. More broadly, this is consistent with prior work showing that anonymizing human regions can affect downstream utility beyond the directly protected content, including object-detection performance on other scene elements~\cite{abdulaziz2025visualprivacy}. However, the variation is noticeably smaller than for worker retention, reinforcing that the worker channel remains the primary point of sensitivity under privacy transformation. Thus, worker obfuscation perturbs both sides of the worker--load relation, but the dominant utility loss still arises from reduced worker detectability.

\paragraph{Human-grounded evaluation sharpens the interpretation of the benchmark.}
The results against human annotations confirm that the main privacy conclusion is not merely an artifact of agreement with the raw pipeline. In particular, cartooning slightly outperforms the raw condition against human labels ($F_2=0.767$ vs.\ 0.757), while Canny-edge, despite having the highest raw-reference stability, does not achieve the strongest agreement with human annotations. This indicates that higher agreement with the raw detector is not itself the target objective. What matters is whether the obfuscated representation preserves the cues needed for human-aligned hazard recognition. At the same time, the absolute performance of the raw proxy remains modest, and Appendix~\ref{app:gt_eval_details} shows that it is recall-oriented but precision-limited. We therefore interpret the current system as a lightweight relational proxy for privacy evaluation rather than as a production-ready suspended-load detector.

\subsection{Limitations and Future Work}
\label{sec:limitations}

\paragraph{Privacy evaluation.}
Our study evaluates \emph{utility under privacy-motivated obfuscation}, but does not directly measure \emph{privacy risk}, such as re-identification or attribute inference. We also consider only one parameterization per obfuscation family. Future work should therefore quantify privacy risk explicitly and characterize the privacy--utility frontier across obfuscation strengths~\cite{deconinck2024visual,hukkelas2023deepprivacy2,kaabachi2025scoping}.

\paragraph{Suspended-load reasoning.}
The hazard detector is a lightweight \emph{relational proxy}, not a calibrated physical risk estimator. It operates on monocular 2D detections and remains sensitive to occlusion, small workers, and perspective distortion. In addition, the current system relies on open-vocabulary detection rather than task-specific lifted-load supervision, so performance depends on prompt design and is constrained by the scarcity of labeled suspended-load data. This is reflected in the absolute raw-video performance of the proxy, which is recall-oriented but precision-limited rather than production-ready. Representative failure cases are shown in Appendix~\ref{app:failure_cases}. Future work should combine dedicated airborne-load annotation with stronger supervision and richer geometric reasoning, including camera calibration, homography, and depth-aware worker--load estimation~\cite{yang2024depthanything}.

\paragraph{Real-world validation and benchmark scale.}
We do not report evaluation on real construction CCTV. This is a central motivation for \textsf{SynthSite}: the real footage available to us did not provide enough positive suspended-load events or controlled diversity for benchmark construction. We therefore use synthetic data to construct a targeted benchmark for this hazard, rather than to claim that the benchmark fully captures the distribution of operational surveillance footage. The present study should accordingly be interpreted as a controlled synthetic benchmark study rather than as a claim of deployment readiness. Although the observed trends are encouraging within the 55 clips of \textsf{SynthSite}, larger-scale validation remains necessary. Future work should expand the benchmark and evaluate on operational CCTV footage with representative suspended-load events.

\paragraph{Beyond worker under suspended load.}
An important next step is to test whether the same privacy--utility pattern holds beyond the current hazard. This includes other \emph{relational} events, such as worker--machinery proximity and geofencing violations, as well as finer-grained worker-centric tasks such as PPE monitoring. In particular, cartooning is promising because it suppresses identity-bearing texture while preserving coarse silhouette and limb structure, but it remains unclear whether this is sufficient for head- and torso-level PPE localization, where the cues are smaller and more appearance-dependent.
\section{Conclusion}
In this work, we introduced \textsf{SynthSite}, a focused synthetic video benchmark of 55 clips for the safety-critical hazard of \emph{worker under suspended load}, together with a privacy-aware hybrid workflow for constructing shareable benchmark data under trust-boundary and deployment constraints. This provides a practical route to benchmarking rare hazards that are difficult, unsafe, or infeasible to capture and release from real construction surveillance footage.

Our central finding is that downstream utility in this setting depends less on preserving worker appearance than on preserving the geometric cues required for relational hazard reasoning. Structure-preserving obfuscations retain substantially more hazard-recognition utility than appearance-smoothing baselines, and agreement with the raw reference is not identical to agreement with human hazard annotations. We also find that worker obfuscation can perturb the load channel even when the load itself is not transformed, suggesting that relational hazard detection can be sensitive to scene context beyond the directly obfuscated region.

Taken together, these results motivate geometry-aware and human-grounded evaluation of privacy-preserving video analytics for construction safety. More broadly, \textsf{SynthSite} is best viewed as a focused starting point rather than a finished benchmark. Expanding benchmark scale, validating on operational CCTV, and extending the framework to other relational construction hazards remain important next steps. We hope this work helps establish privacy-aware benchmarking as a practical foundation for safety-critical construction video analytics.

\section*{Acknowledgements}
We thank Prof. Roy Ka-Wei Lee at the Singapore University of Technology and Design (SUTD) for his valuable feedback and guidance in improving this work.

{
    \small
    \bibliographystyle{ieeenat_fullname}
    \bibliography{main}
}

\clearpage
\appendix
\clearpage
\appendix
\setcounter{page}{1}

\maketitlesupplementary

\section{Generation Workflow Details}
\label{app:generation_details}

\subsection{Model Allocation Across Environments}
\label{app:model_allocation}

The workflow instantiated in this paper uses Qwen2.5-VL-7B~\cite{qwenteam2025qwen25vl} inside the sensitive environment to convert raw frames into approved textual intermediaries. In the non-sensitive environment, Nano Banana 2~\cite{google2026nanobanana} is used to generate synthetic seed images and Veo 3.1 Fast~\cite{google2025veo} is used to generate publicly shareable synthetic videos. These videos are then used to adapt Wan2.2-I2V-A14B~\cite{wanteam2025wan} via LoRA~\cite{hu2021lora} (details are provided in Table~\ref{tab:lora_settings}), and the resulting adapter is imported back into the sensitive environment for internally controlled generation.

Model allocation was driven by deployment constraints rather than a fixed preference for open or closed models. The local vision--language model and local video generator were selected because their weights permitted deployment inside the trusted environment, while the external image and video generators were used because they provided stronger prompt adherence and temporal consistency for proxy-data creation. On our available hardware, an RTX~5090 (32\,GB VRAM) supported inference but not stable LoRA adaptation of the target video model, so LoRA training was performed on an RTX~6000 Pro (96\,GB VRAM) using 8 curated publicly shareable synthetic videos.

\begin{table}[t]
\centering
\caption{LoRA fine-tuning settings for Wan2.2-I2V-A14B trained on 8 curated publicly shareable synthetic videos using AI-Toolkit.}
\label{tab:lora_settings}
\small
\setlength{\tabcolsep}{4pt}
\begin{tabular}{ll}
\toprule
\textbf{Parameter} & \textbf{Value} \\
\midrule
Base Model                    & Wan2.2-I2V-A14B~\cite{wanteam2025wan} \\
LoRA Network Module           & Linear \& Convolutional Adaptation \\
Rank ($r$)                    & 16 \\
Alpha ($\alpha$)              & 16 \\
Learning Rate                 & $1 \times 10^{-4}$ \\
Optimizer                     & AdamW 8-bit \\
Training Steps                & 3{,}000 \\
Input Resolution              & $768$ pixels \\
Loss Function                 & MSE \\
Mixed Precision               & BF16 \\
Target Layers                 & UNet only (text encoder frozen) \\
Batch Size per GPU            & 1 \\
Gradient Accumulation         & 1 \\
Noise Scheduler               & FlowMatch \\
Quantisation                  & qfloat8 (UNet \& text encoder) \\
Training Framework            & AI-Toolkit by Ostris~\cite{ostris2024aitoolkit} \\
\bottomrule
\end{tabular}
\end{table}


\begin{table*}[t]
\centering
\caption{Representative sanitized textual intermediary used in the generation workflow. The example preserves hazard-relevant structure, scene layout, and surveillance context while omitting site-identifying details.}
\label{tab:text_intermediary_example}
\small
\setlength{\tabcolsep}{6pt}
\renewcommand{\arraystretch}{1.12}
\begin{tabularx}{\textwidth}{p{0.18\textwidth}X}
\toprule
\textbf{Component} & \textbf{Content} \\
\midrule

\textbf{System Prompt} &
\textbf{Task:} You are an expert construction-site image analyst. Produce a detailed but privacy-aware textual description of the provided construction-site image. The description will be used for synthetic regeneration of a hazard-relevant scene, so it should preserve spatial layout, worker--machinery relations, construction context, and surveillance characteristics while avoiding direct reproduction of site-identifying details.

You MUST describe the following if visible: camera viewpoint and framing; lighting, weather, and time-of-day category; number and relative positions of workers; worker posture, activity, and PPE category; machinery type, position, and operational state; suspended materials and their relation to workers; construction stage, structural elements, and visible materials; ground conditions; safety barriers and temporary structures; occlusion, clutter, scale variation, and other nuisance factors relevant to hazard interpretation.

You MUST represent background context only at an abstract level, for example dense tropical urban surroundings, elevated roadway or rail infrastructure, roadside vegetation, adjacent residential or commercial buildings, and multilingual urban signage without reproducing exact text.

You MUST NOT include exact timestamps, block numbers, street names, company names, machine brand names, readable signage text, license plates, unique landmarks, or any other site-identifying details unless they have been explicitly approved.

\textbf{Output format:} A single continuous paragraph of comma-separated descriptive phrases, written foreground to background. Focus on geometry, spatial relations, operational context, and surveillance realism rather than exact identity-bearing details.
\\[0.5em]

\textbf{User Prompt} &
Analyze this construction-site image and produce a privacy-aware reverse-prompt description suitable for synthetic regeneration. Emphasize worker positions relative to suspended loads, machinery, structures, and ground conditions, together with camera viewpoint, lighting, clutter, occlusion, and other factors that affect relational hazard reasoning. Preserve hazard-relevant structure, but avoid site-identifying details and exact textual content visible in the scene.
\\[0.5em]

\textbf{Model Output} &
Elevated wide-angle CCTV view of an outdoor urban construction site in bright late-afternoon light, early-stage earthworks with exposed soil, temporary fencing, and scattered construction materials, multiple workers distributed across the foreground and midground wearing hard hats and reflective safety vests, one worker positioned near a tracked excavator, another near the base of a stationary crane, suspended construction material visible across the central work zone with clear worker--equipment spatial relationships, temporary utility structures at the far right, moderate clutter and partial occlusion, adjacent roadway and dense mid-rise urban buildings in the background, roadside vegetation extending into the upper frame, surveillance-style perspective with wide field of view and realistic construction-site layout.
\\

\bottomrule
\end{tabularx}
\end{table*}


\subsection{Textual Intermediaries}
\label{app:reverse_prompt_examples}

To generate synthetic construction-site imagery without directly exposing sensitive footage, we use a two-stage text-intermediary pipeline. In the first stage, real CCTV frames are processed inside the sensitive environment by a locally hosted \texttt{Qwen2.5-VL-7B-Instruct} vision--language model~\cite{qwenteam2025qwen25vl}, which produces structured, privacy-aware scene descriptions covering camera viewpoint, worker locations and poses, machinery type and operating context, suspended materials, structural elements, ground conditions, and other hazard-relevant cues. Because inference is performed locally, no raw site imagery is transmitted to external services.

The resulting description is then used as input to \texttt{Nano Banana 2}~\cite{google2026nanobanana} to synthesize a novel seed image. This seed image is intended to preserve coarse scene layout and hazard-relevant structure while reducing direct exposure to the original footage. In practice, the textual intermediaries emphasize worker--load relations, suspension context, scene layout, and nuisance factors such as occlusion, clutter, and lighting, while avoiding site-identifying details and direct reconstruction of sensitive visual content.

For privacy reasons, the prompt template shown in Table~\ref{tab:text_intermediary_example}) is a sanitized but structurally representative example: it preserves the hazard-relevant fields used for downstream synthetic generation while omitting site-identifying details and exact scene identifiers. 


\begin{figure*}[!t]
  \centering
  \includegraphics[width=\linewidth]{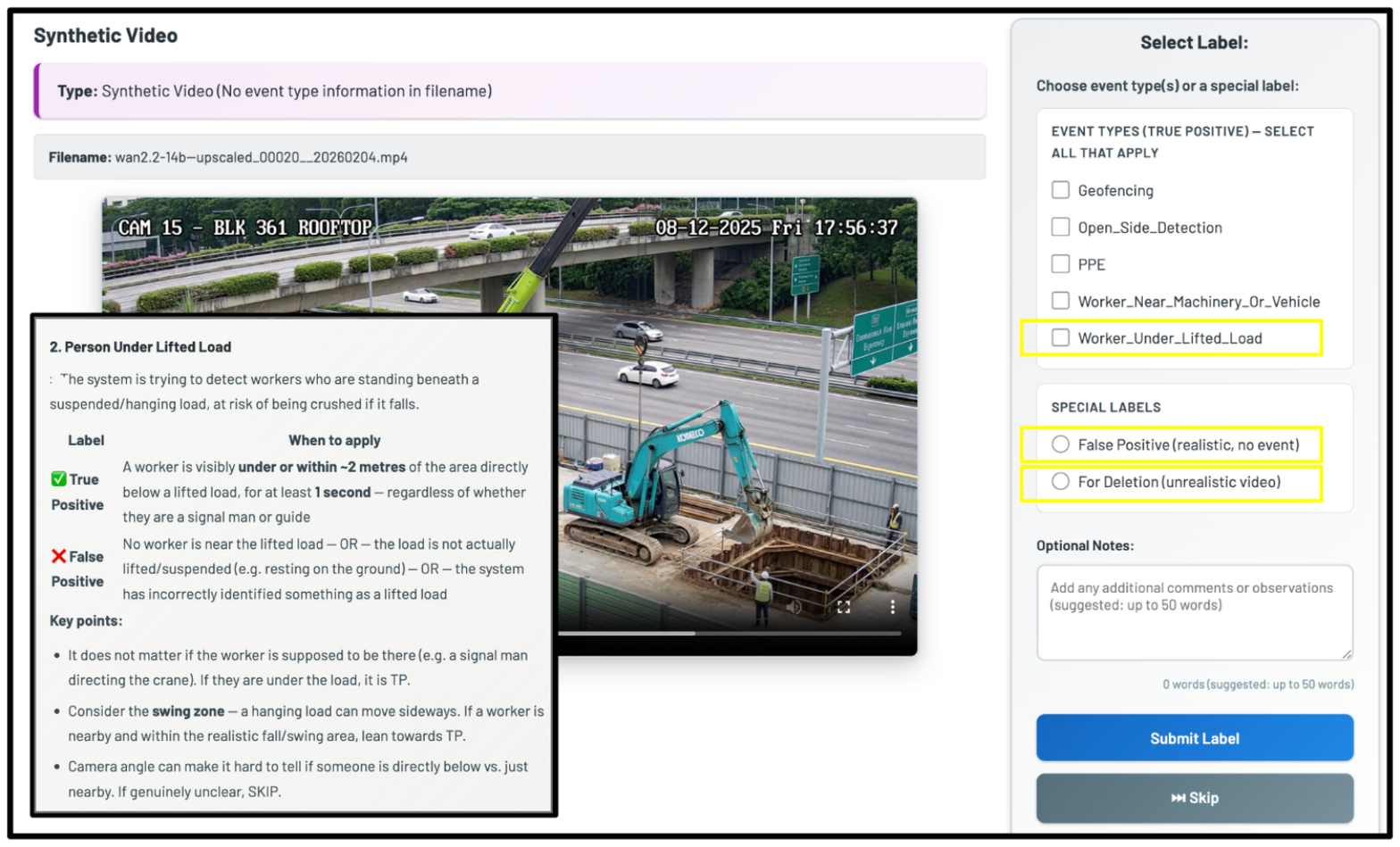}
\caption{\textbf{Custom annotation platform used for \textsf{SynthSite} labeling.} Annotators reviewed each clip together with an inline event-specific rubric defining unsafe and safe cases, hard negatives, and exclusion criteria for ambiguous or low-quality clips. Although the platform supported multiple construction-safety event types, this paper uses only the worker-under-suspended-load rubric. In the interface, \textit{True Positive} corresponds to \emph{unsafe}, \textit{False Positive (realistic, no event)} corresponds to \emph{safe}, and \textit{For Deletion} marks clips excluded from the benchmark due to ambiguity or quality failures.}
  \label{fig:video_annotation_platform}
\end{figure*}


\section{Dataset Annotation Platform and Rubric}
\label{app:rubric}

Figure~\ref{fig:video_annotation_platform} shows the custom web-based annotation platform used for rubric-guided labeling. Annotators reviewed each clip together with an inline event-specific rubric, so that decisions were made against a consistent operational definition rather than subjective judgment. The platform supported multiple construction-safety event types, but this paper uses only the \textit{worker under suspended load} rubric. The interface also allowed ambiguous clips to be skipped and low-quality clips to be flagged for exclusion.

For this paper, a clip was labeled \emph{unsafe} when a visibly suspended load was present and a worker remained within the fall zone beneath the load for at least 1 second. Clips were labeled \emph{safe} when no worker was present in the fall zone, the load was not actually suspended, or the scene created a misleading impression of suspension due to perspective or occlusion. Consistent with the rubric shown in Figure~\ref{fig:video_annotation_platform}, annotators were instructed that authorized role does not override the safety label: if a worker is under the load, the clip is still considered unsafe. Ambiguous cases were excluded rather than forced into a label.

The platform additionally distinguished between valid negative clips and clips unsuitable for benchmarking. Clips depicting realistic scenes with no safety event were retained as negative examples, while clips with severe synthetic artefacts or other quality failures were flagged for deletion and excluded from the final benchmark. In the interface, these were presented using generic platform labels such as \textit{True Positive}, \textit{False Positive}, and \textit{For Deletion}; in the final \textsf{SynthSite} benchmark, these map to the paper's \emph{unsafe}, \emph{safe}, and excluded categories, respectively.

Consistent with the main paper, clips were independently reviewed by two annotators from a pool of eight, and disagreements were resolved by a third annotator through adjudication.

\section{Implementation Details for Worker Obfuscation}
\label{app:obfuscation_details}

Worker segmentation masks are obtained using a shared Ultralytics YOLO segmentation model (\texttt{yolo26x-seg.pt}) restricted to the \texttt{person} class. To improve coverage of small workers in wide surveillance frames, segmentation is run with tiled inference (\texttt{InferenceSlicer}, slice size $640\times640$, overlap $160\times160$) at confidence threshold $0.1$, and detections are associated across frames with ByteTrack (\texttt{lost\_track\_buffer}=60). Mask dilation was disabled in the final pipeline. The same per-frame masks are reused across all obfuscation conditions so that performance differences arise from the transformation itself rather than from differences in mask prediction.

The evaluated transformations are applied as follows. \textbf{Canny-edge} converts the frame to grayscale, applies Canny edge detection with thresholds $(100, 200)$, and replaces the union of all worker-mask regions with the resulting edge map. \textbf{Cartooning} applies $K$-means color quantization with $K=4$, followed by bilateral filtering with parameters \texttt{(9, 75, 75)} for spatial and color smoothing. \textbf{Pixelation} downsamples each worker crop to a square grid of size $\max(1, \min(16, \lfloor w/4 \rfloor))$, where $w$ is the crop width in pixels, using bilinear interpolation for downsampling and nearest-neighbor interpolation for upsampling. \textbf{Blur} applies Gaussian smoothing with kernel size $(51, 51)$ and $\sigma=0$. For pixelation and blur, a white contour of thickness 2 pixels is drawn around each worker instance to preserve visible extent and instance separability for downstream relational localization.


\begin{table*}[h]
\centering
\caption{Pipeline parameters of the relational hazard detection system.}
\label{tab:supp_params}
\small
\setlength{\tabcolsep}{4pt}
\begin{tabular}{llp{7cm}}
\toprule
\textbf{Parameter} & \textbf{Default} & \textbf{Purpose} \\
\midrule
\texttt{YOLO\_CONF}                  & 0.15  & Detection confidence threshold \\
\texttt{YOLO\_IOU}                   & 0.40  & NMS IoU threshold \\
\texttt{SUSPENDED\_CLEARANCE\_RATIO} & 0.12  & Minimum clearance to classify load as suspended \\
\texttt{NEARBY\_WORKER\_X\_EXPANSION}& 1.25  & Horizontal search radius ($\times$ load width) \\
\texttt{ASSUMED\_WORKER\_HEIGHT\_M}  & 1.75  & Assumed worker height for scale normalization (m) \\
\texttt{V\_DRIFT\_MIN}               & 0.4   & Minimum horizontal velocity floor (m/s) \\
\texttt{VELOCITY\_BUFFER\_SIZE}      & 8     & Rolling buffer size for velocity estimation (frames) \\
\texttt{MIN\_HAZARD\_EXPANSION\_PX}  & 30    & Minimum zone expansion at load level (px) \\
\texttt{VERTICAL\_GATE\_PIXELS}      & 40    & Vertical compatibility gate (px) \\
\texttt{WINDOW\_SECONDS}             & 1.0   & Temporal sliding window duration (s) \\
\texttt{K\_THRESHOLD}                & 10    & Positive frames required to trigger \textsc{Unsafe} \\
\texttt{MAX\_NEGATIVE\_GAP}          & 2     & Maximum gap length for temporal infilling (frames) \\
\bottomrule
\end{tabular}
\end{table*}


\section{Relational Hazard Pipeline Details}
\label{app:relational_hazard_pipeline_details}

This section describes the full implementation of the relational hazard detection pipeline introduced in Section~\ref{sec:relational_hazard_pipeline} of the main paper. The pipeline parameters are provided in Table~\ref{tab:supp_params}.

\subsection{Object Detection and Tracking}

We use YOLO-World (\texttt{yolov8x-worldv2}) with two task-critical open-vocabulary prompts: \texttt{person} and \texttt{hoisted load with crane hook}. Detection operates at a confidence threshold of 0.15 and an NMS IoU threshold of 0.40 to favor recall in this safety-oriented setting. ByteTrack associates detections across frames, enabling temporal smoothing and per-load motion estimation.

\subsection{Suspension Inference: Workers as Ground-Level Proxies}

Since no fixed ground plane is available across clips, we use worker footpoints---defined as the bottom-center of each person bounding box---as local ground-level references. For each detected load, candidate nearby workers are those whose footpoint x-coordinate lies within $1.25\,w_\text{load}$ of the load center, where $w_\text{load}$ is the load bounding-box width. This proximity filter restricts ground references to workers at a plausibly similar depth.

For each nearby worker $i$, we compute the normalized clearance ratio
\begin{equation}
r_i^\text{clr} = \frac{y_i^\text{foot} - y^\text{bottom}_\text{load}}{h_i^\text{worker}},
\end{equation}
where $y_i^\text{foot}$ is the worker footpoint y-coordinate, $y^\text{bottom}_\text{load}$ is the load bounding-box bottom edge, and $h_i^\text{worker}$ is the worker bounding-box height in pixels. A positive clearance ratio indicates that the load is elevated above the local ground reference. A load is classified as \textit{suspended} when the maximum clearance ratio among all nearby workers exceeds a threshold of 0.12. If no nearby workers are detected, the pipeline does not infer suspension from this proxy and treats the load as non-suspended for downstream hazard evaluation.

\subsection{Hazard-Zone Geometry}

\paragraph{Scale estimation.}
In the absence of camera intrinsics, a rough per-frame pixel-to-meter scale proxy is estimated using detected person heights:
\begin{equation}
\rho = \frac{\tilde{h}^\text{worker}_\text{px}}{H_\text{worker}},
\end{equation}
where $\tilde{h}^\text{worker}_\text{px}$ is the median detected worker height in pixels across the frame, $H_\text{worker}=1.75$~m is the assumed real-world worker height, and $\rho$ is the resulting scale estimate. This proxy is approximate and is used primarily to normalize hazard-zone expansion across clips with different apparent scales.

\paragraph{Load height above ground.}
The load height in meters is estimated as
\begin{equation}
h_\text{load} = \frac{y^\text{foot}_\text{best} - y^\text{bottom}_\text{load}}{\rho},
\end{equation}
where $y^\text{foot}_\text{best}$ is the footpoint of the nearby worker with the largest clearance ratio.

\paragraph{Trapezoidal fall-zone proxy.}
A falling load may undergo lateral displacement proportional to its horizontal velocity and free-fall duration. The fall time is estimated from basic kinematics:
\begin{equation}
t_\text{fall} = \sqrt{\frac{2\,h_\text{load}}{g}},
\end{equation}
where $g$ denotes gravitational acceleration. Let $\hat{v}_x$ denote the estimated horizontal load velocity computed from tracked load motion over a rolling buffer of 8 frames. The effective horizontal velocity is then
\begin{equation}
v_\text{eff} = \max\!\left(v_\text{drift},\; \hat{v}_x\right),
\end{equation}
where $v_\text{drift}=0.4$~m/s is a minimum drift floor introduced to account for wind and residual crane swing. The maximum lateral displacement is
\begin{equation}
\delta_\text{lat} = v_\text{eff}\, t_\text{fall}.
\end{equation}

The fall-zone proxy is modeled as a trapezoid centered on the load's horizontal position. Its half-width expands linearly from the load bottom to the estimated ground level beneath the load, ranging from $w_\text{load}/2 + \epsilon$ at the load bottom to $w_\text{load}/2 + \delta_\text{lat}$ at the ground, where $\epsilon=30$~px is a minimum expansion floor. At any worker foot y-coordinate $y^\text{foot}$, the zone half-width is computed by linear interpolation between these two bounds.

\subsection{Worker-in-Zone Test}

A worker is flagged as exposed if two conditions hold simultaneously: (i) the worker footpoint x-coordinate falls within the interpolated zone half-width at the worker foot y-coordinate, and (ii) the worker footpoint lies at least 40~px below the load bottom edge. The second condition serves as a vertical compatibility gate that suppresses false positives from workers detected on elevated structures above the load.

\subsection{Temporal Smoothing and Clip-Level Decision}

Per-frame hazard indicators are aggregated over a 1-second sliding window (15 frames at 15~FPS). Short negative gaps of up to 2 consecutive frames surrounded by positive detections on both sides are filled to compensate for transient detection dropouts. A clip is labeled \textsc{Unsafe} when at least 10 positive frames occur within the window ($\approx 67\%$ occupancy), and this state is latched for the remainder of the clip. This temporal rule is intentionally more tolerant than the annotation rule in order to remain robust to intermittent detector failures. A summary of all pipeline parameters is provided in Table~\ref{tab:supp_params}.

\section{Detector Selection and Prompt Design}
\label{app:detector_selection}

Because labeled suspended-load data is scarce, we used an open-vocabulary detector rather than training a task-specific model. In a small pilot on raw-video hazard analysis, we compared YOLO-World (\texttt{yolov8x-worldv2}) and the Grounding DINO Base model. YOLO-World was more suitable for our pipeline for two reasons. First, it produced more usable lifted-load detections in practice, whereas Grounding DINO often either missed the load entirely or assigned the load label to large crane regions. Second, it was substantially faster in our setting: on the same RTX-5090 hardware, Grounding DINO required more than 15 minutes to process all 55 videos, whereas YOLO-World required less than 2 minutes.

For YOLO-World, we found the most stable performance when prompting both the target classes and common construction-machinery distractors. The final prompt set was:
\begin{verbatim}
person
hoisted load with crane hook
excavators
trucks
Tower crane
Truck crane
Roller
Bulldozer
Excavator
Truck
Loader
Pump truck
Concrete truck
Pile truck
\end{verbatim}

Including common machinery distractors improved load localization stability relative to more granular alternatives such as prompting \texttt{hook} or \texttt{load} separately, which were more prone to confusion with crane structures and background machinery.

For reference, the raw-video Grounding DINO baseline achieved 61.8\% accuracy and $F_2=0.625$ on clip-level hazard recognition, below the higher-recall YOLO-World-based proxy used in the main experiments. We therefore selected YOLO-World as the detector backbone for the privacy evaluation.

We also did not pursue extensive detector fine-tuning. Preliminary attempts suggested unstable multi-class behavior under limited labels, including degraded person detection, and detector optimization was outside the scope of this privacy-focused study.

\begin{table}[t]
\centering
\caption{Clip-level precision, recall, and $F_2$ against human safe/unsafe annotations. The \emph{unsafe} class is treated as positive. The table exposes the precision--recall trade-off underlying the main-paper $F_2$ results.}
\label{tab:gt_eval_details}
\small
\setlength{\tabcolsep}{5pt}
\begin{tabular}{lccc}
\toprule
\textbf{Method} & \textbf{Prec $\uparrow$} & \textbf{Rec $\uparrow$} & \textbf{$F_2 \uparrow$} \\
\midrule
RAW         & 0.523 & 0.852 & 0.757 \\
Canny-edge  & 0.489 & 0.852 & 0.742 \\
Cartooning  & \textbf{0.548} & 0.852 & \textbf{0.767} \\
Pixelation  & 0.535 & 0.852 & 0.762 \\
Blur        & 0.488 & 0.741 & 0.671 \\
\bottomrule
\end{tabular}
\end{table}



\begin{figure*}[t]
  \centering
  \includegraphics[width=\linewidth]{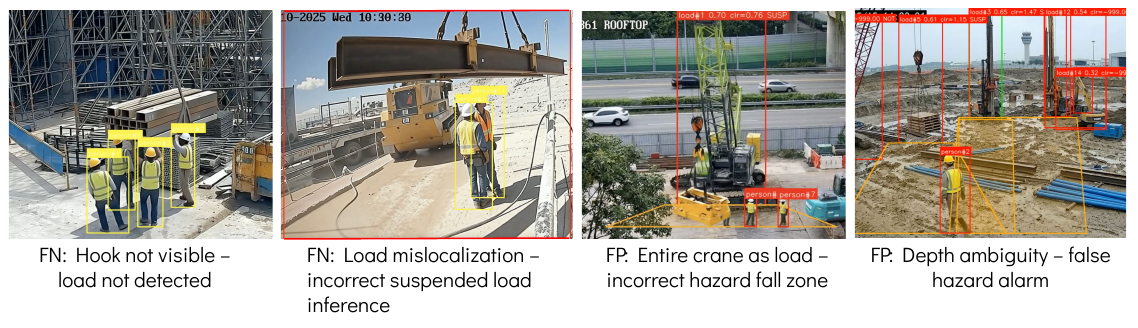}
    \caption{\textbf{Representative failure cases of the relational hazard pipeline.} From left to right: (\textit{FN}) hook not visible, so the suspended load is not detected; (\textit{FN}) load mislocalization, where an incorrect bounding box suppresses suspension inference by placing the load bottom too close to nearby workers; (\textit{FP}) entire-crane-as-load confusion, which produces an inflated fall-zone proxy; and (\textit{FP}) depth ambiguity, where monocular 2D overlap triggers a false hazard despite true depth separation. These examples suggest that many residual errors arise from the load channel and monocular geometric ambiguity rather than from temporal aggregation alone.}
  \label{fig:failure_cases}
\end{figure*}

\section{Additional Results and Failure Analysis}
\label{app:additional_results_failure_analysis}

\subsection{Agreement with Human Annotations: Precision--Recall Breakdown}
\label{app:gt_eval_details}

Table~\ref{tab:gt_eval_details} reports clip-level precision, recall, and $F_2$ against the human safe/unsafe annotations, treating \emph{unsafe} as the positive class. We emphasize $F_2$ in the main paper because the task is safety-oriented and therefore recall-sensitive, but provide the precision--recall breakdown here to make the underlying error trade-offs explicit.

Across RAW, Canny-edge, cartooning, and pixelation, recall remains fixed at 0.852, indicating that the main differences among these conditions arise from precision rather than missed unsafe events. Cartooning achieves the strongest $F_2$ against human annotations because it improves precision while preserving recall, whereas blur degrades both precision and recall.


\subsection{Representative Failure Cases of Relational Hazard Reasoning}
\label{app:failure_cases}

Figure~\ref{fig:failure_cases} shows representative failure cases from the relational hazard pipeline. These examples are not intended as an exhaustive taxonomy, but they illustrate the dominant residual error modes observed during qualitative inspection.

A recurring pattern is that failures arise more often from the \emph{load channel} than from temporal aggregation itself. In particular, we observe missed suspended-load detections when hook context is weak or absent, load mislocalization that suppresses suspension inference by shifting the estimated load bottom, entire-crane-as-load confusion that inflates the fall-zone proxy, and depth ambiguity where monocular 2D overlap produces a false hazard despite true depth separation.

These cases reinforce the interpretation in the main paper: the current system should be viewed as a lightweight relational proxy for privacy-robustness analysis rather than as a production-ready suspended-load detector. Future gains are therefore more likely to come from stronger lifted-load detection, dedicated airborne-load supervision, and depth-aware reasoning than from further tuning of the temporal decision rule alone.

\end{document}